\documentclass[letterpaper, 10 pt, conference]{ieeeconf}  

\IEEEoverridecommandlockouts                              
\usepackage{slashbox}
\usepackage{graphicx}
\usepackage{textgreek}
\usepackage{amsmath}
\usepackage[font=footnotesize]{caption}
\usepackage{subcaption}
\usepackage{booktabs}
\usepackage{multirow}
\usepackage{array}
\usepackage{cite}
\usepackage{mathtools}
\usepackage{comment}
\usepackage{amsmath}
\usepackage{amsfonts}
\usepackage{amssymb}
\usepackage[hyphens]{url}
\usepackage{xcolor}
\usepackage{hyperref}
\usepackage[linesnumbered,ruled,vlined]{algorithm2e}

\SetCommentSty{mycommfont}

\SetKwInput{KwInput}{Input}                
\SetKwInput{KwOutput}{Output}              

\title{\LARGE \bf
MuCaSLAM: CNN-Based Frame Quality Assessment for Mobile Robot with Omnidirectional Visual SLAM
}

\author{Pavel Karpyshev, Evgeny Kruzhkov, Evgeny Yudin, Alena Savinykh, Andrei Potapov, Mikhail Kurenkov,\\
Anton Kolomeytsev, Ivan Kalinov, and Dzmitry Tsetserukou
\thanks{All authors are with the Intelligent Space Robotics Laboratory, Center for Digital Engineering, Skolkovo Institute of Science and Technology, Moscow, Russian Federation.
{\tt \{pavel.karpyshev, evgeny.kruzhkov, evgeny.yudin, alena.savinykh, andrei.potapov, mikhail.kurenkov, anton.kolomeytsev, d.tsetserukou\}@skoltech.ru, ivan.kalinov@skolkovotech.ru}}%
}

\begin{document}
\maketitle
\thispagestyle{empty}
\pagestyle{empty}

\begin{abstract}

In the proposed study, we describe an approach to improving the computational efficiency and robustness of visual SLAM algorithms on mobile robots with multiple cameras and limited computational power by implementing an intermediate layer between the cameras and the SLAM pipeline. In this layer, the images are classified using a ResNet18-based neural network regarding their applicability to the robot localization. The network is trained on a six-camera dataset collected in the campus of the Skolkovo Institute of Science and Technology (Skoltech). For training, we use the images and ORB features that were successfully matched with subsequent frame of the same camera (“good” keypoints or features). The results have shown that the network is able to accurately determine the optimal images for ORB-SLAM2, and implementing the proposed approach in the SLAM pipeline can help significantly increase the number of images the SLAM algorithm can localize on, and improve the overall robustness of visual SLAM. The experiments on operation time state that the proposed approach is at least 6 times faster compared to using ORB extractor and feature matcher when operated on CPU, and more than 30 times faster when run on GPU. The network evaluation has shown at least 90\% accuracy in recognizing images with a big number of “good” ORB keypoints. The use of the proposed approach allowed to maintain a high number of features throughout the dataset by robustly switching from cameras with feature-poor streams.
\end{abstract}

\section{Introduction}

\begin{figure} [!t] 
\begin{center}
\includegraphics[width=8.4 cm]{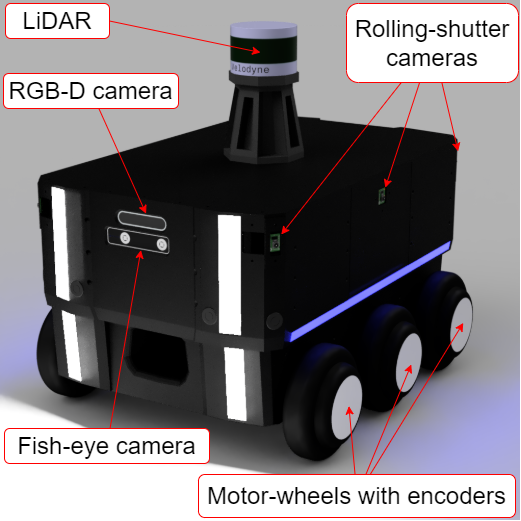}
\caption{Render and hardware equipment of autonomous delivery robot HermesBot.}
\vspace{-1.5em}
\label{hermes_render}
\end{center}
\end{figure}

\subsection{Motivation}
The area of e-commerce is growing at an astonishing rate, and is predicted to grow up to more than EUR 2.5 trillion by the year 2023, according to \cite{tmirob}. This growth has inevitably led to advances in all surrounding areas, including the market of goods delivery. And in that area, one of the most important parts is the so-called “last-mile delivery”, that is, the coverage of the distance between the nearest shop or warehouse directly to the customer. This area in particular directly influences customer satisfaction, which is essential for the success of logistic companies. Nowadays, this kind of delivery is typically performed by foot couriers, bikes, or vehicles. However, recent studies have shown that due to the traffic congestion in big cities, the delivery time using vehicles has significantly increased. 


The solution to the problem of automatized last-mile delivery would be the use of autonomous ground vehicles (AGVs). Such solutions have already proven effective and cost-efficient, and have created a huge demand on the market\cite{forbes2020rovers}. For example, the company Yandex has already reached the final testing phase of their own delivery rover\cite{yandexrover}. Apart from Yandex, multiple huge companies and startups, e.g., Starship Technologies, Postmates, Ford, Amazon, etc., are developing or already testing their ground robotic delivery solutions.



\subsection{Problem Statement}

\begin{figure} [!t] 
\begin{center}
\includegraphics[width=8.4 cm]{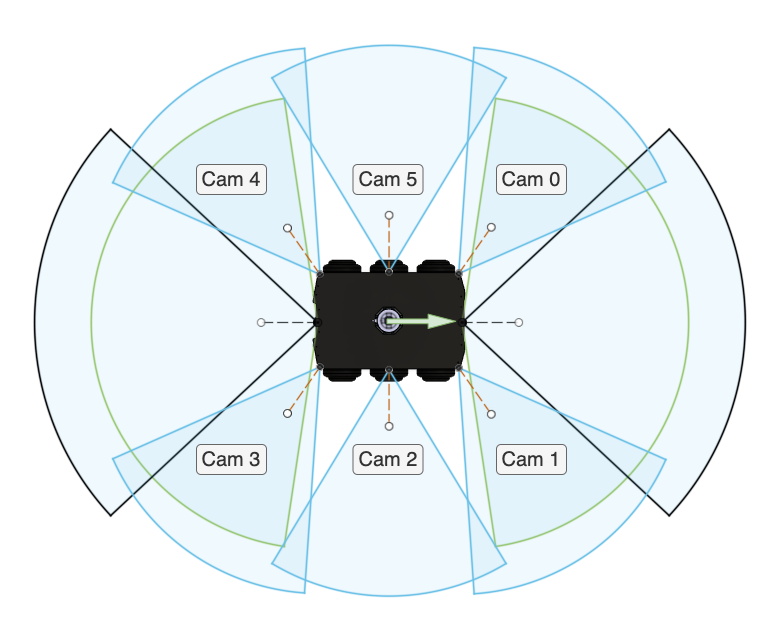}
\caption{Field of view (FoV) of HermesBot cameras. Blue sectors, green sectors, and black sectors depict the FoV of RasPi V2 cameras, RealSense T265, and Intel RealSense D435, respectively.}
\vspace{-2.5em}
\label{fig:hermes_fov}
\end{center}
\end{figure}

Localization and mapping algorithms are an essential part of every autonomous robot, and are often combined into one module – Simultaneous Localization and Mapping (SLAM). State-of-the-art SLAM algorithms are able to utilize a wide variety of sensors, from inexpensive rolling-shutter cameras to LiDARs and 3D cameras of different types. The AGVs that are currently being developed or tested mostly utilize LiDARs for the SLAM task, and that is dictated by the robustness and versatility of these sensors: LiDARs are able to operate regardless of weather and light conditions, and provide the information on the robot's surroundings in 360 degrees. Contrariwise, these sensors provide a relatively sparse point cloud of the area, and are highly suboptimal for the use in other systems of the robot, such as object and human detection. Moreover, these sensors are highly expensive and hard to manufacture.

The Visual SLAM algorithms utilize the information from visual sensors, e.g., global- and rolling-shutter cameras, for the localization and mapping. Such approaches use the data from much cheaper sensors, and thus, make the systems based on them more cost-effective. Moreover, the cameras installed on the robot can be used for multiple purposes apart from localization, for example the aforementioned object and human detection, further decreasing the required set of hardware and the overall system cost. However, the use of low-cost visual sensors has its own drawbacks: cameras are much less robust to changes in illumination and weather conditions, and show overall worse results, compared to LiDARs, in the scope of the SLAM problem. Furthermore, the FoV of cameras is limited, thus making such systems even more dependent on the quality of data.

For the operation of visual SLAM algorithms, the data from the cameras must contain diversified information with a high number of peculiarities for visual feature extraction; otherwise it would not be able to localize at all. The robustness of visual SLAM can be increased by utilizing multiple cameras oriented in different directions, yet current state-of-the-art visual SLAM algorithms are computationally complex, and that kind of system would require much more computational power rather than one-camera setup, leading to the trade-off between robustness and cost-effectiveness.

We propose an approach that would increase the robustness of visual SLAM without a significant increase in computational complexity for a robot with omnidirectional vision. This is achieved using a lightweight neural network capable of estimating the quality of visual data from each camera in a multi-camera setup. After the quality is estimated, the system only uses the data from the cameras with the best data quality, thus making it unnecessary to use the computationally complex SLAM methods on all cameras of the system.


\begin{figure*} [!t] 
\begin{center}
\includegraphics[width=16.4 cm]{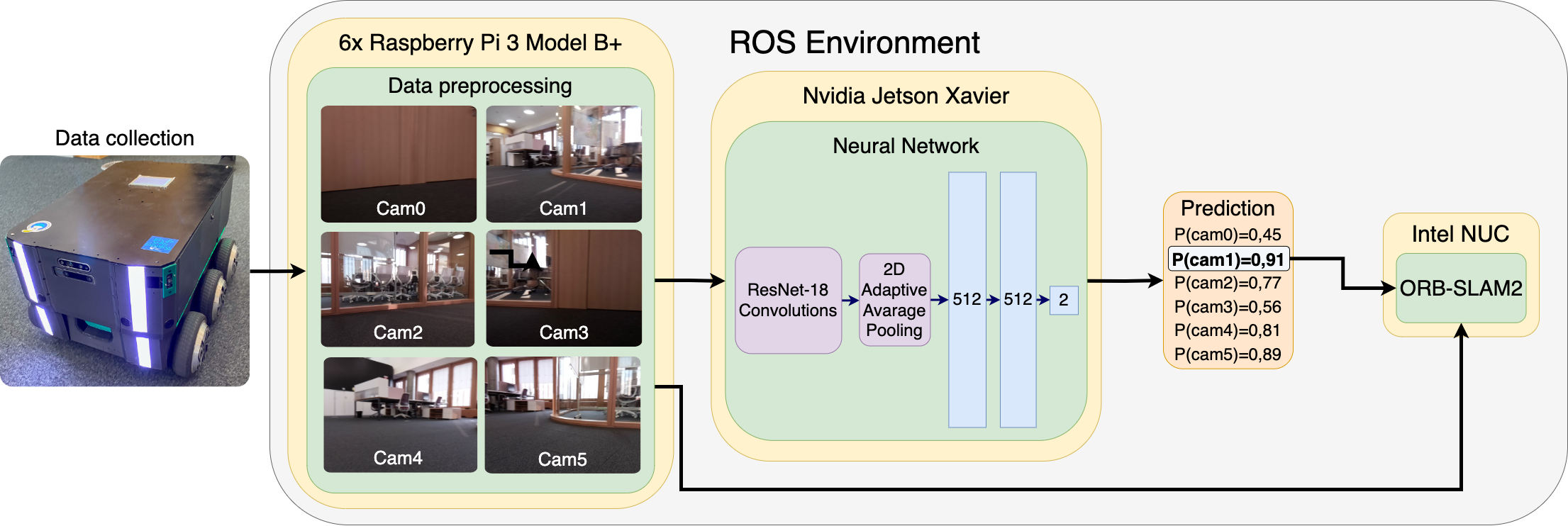}
\caption{The system shares the common ROS environment. Images captured on multiple cameras are preliminarily processed by Raspberry Pi 3 Model B+. Then, the proposed neural network module, launched on Nvidia Jetson Xavier, processes the obtained images as a batch and predicts the best camera in terms of number of features. The ``best" image is transferred to ORB-SLAM2 input, that is launched on Intel NUC, to adapt it for operation in multi-camera setup.}
\vspace{-2.5em}
\label{fig:pipeline_architecture}
\end{center}
\end{figure*}
\subsection{Related Works}
Over the past few years, there has been an evolution of the setups used for SLAM tasks. Particularly, many approaches employing visual sources of information were developed, forming the visual SLAM research area.
VINS-Mono \cite{qin2018vins} is one of the most successful algorithms using a monocular camera along with IMU data. In it, IMU data and feature data are combined using a non-linear optimization method. The next stage in the development of this method is OpenVINS \cite{geneva2020openvins}, that combines various improvements of the VINS-Mono algorithm. At the same time, algorithms were developed that use stereo cameras in SLAM (for example, ORB-SLAM2 \cite{mur2017orb}), which significantly increases the ability to estimate the trajectory, and reduces drift when moving. The next step in improving methods is ElasticFusion \cite{gallagher2018collaborative}, where a solution was presented for reconstructing a scene using multiple independent cameras. The algorithm allows determining the overlap of cameras in space and combine the reconstructions of the scene obtained by them. In the article \cite{kuo2020redesigning}, Kuo et al. made significant changes to the SLAM pipeline, including the addition of an adaptive initialization scheme, a sensor-independent keyframe selection algorithm, and a voxel map. These improvements allow increasing the system robustness by creating complex multi-camera configurations.


Several works aim to extend and improve the performance of visual odometry and SLAM techniques for multi-camera setups. For example, Liu et al. \cite{liu2018towards} described the method based on the combination of pose tracker and local mapper that is directly aimed at minimizing the photometric error. This approach leads to an increase in the accuracy of localization results, even in complex nighttime environments.

AMV-SLAM \cite{yang2021asynchronous} introduces a generalized multi-camera framework for asynchronous sensor observations that groups asynchronous frames during mapping, tracking, and loop closure. This method increases accuracy and robustness in complex real-world conditions, in opposition to classical SLAM that uses synchronized data sources.

The method MultiCol-SLAM \cite{urban2016multicol} shows the improvements of the ORB-SLAM \cite{mur2017orb}. MultiCol-SLAM expands ORB-SLAM to multi-camera SLAM by implementing multi-keyframes, multi-camera loop closure, and made some performance improvements. This method increases the quality and robustness of perception in challenging environments. Moreover, Ye et al. \cite{ye2019robust} proposed a method that estimates the movement of vehicles by using multi-camera systems with IMU for the improvement of the SLAM system robustness.

The article \cite{ventura2015efficient} proposes a method that uses a first-order approximation to relative position estimation to simplify and speed up the performance of the algorithm and improve the accuracy of position estimation. However, such solution can approach real-time conditions only within a random sample. The iterative scheme presented in the article \cite{kneip2014efficient} uses a low-dimensional factorization of the generalized problem of determining the relative position. 

Deep learning is also being extensively used in state-of-the-art SLAM systems. For example, SuperPoint \cite{detone2018superpoint} and SuperGlue \cite{sarlin2020superglue} improve detection, description, and matching of keypoints by using a deep learning approach. Moreover, modern SLAM systems can use objects detected by deep neural network (DNN) as features, such as the CubeSLAM method \cite{yang2019cubeslam}. This technique increases robustness to poor texture environment like indoor with monotonic walls.

Approaches using neural networks for image preprocessing have already been used in other modules of robot operation. For example, Protasov et al. \cite{protasov2021cnn} used the same hardware platform to implement a similar system into the object detection pipeline, that allowed to decrease the computational load on the robot's computing module.

\subsection{Contribution}
We propose and evaluate an approach that increases the robustness and computational efficiency of visual SLAM algorithms in multi-camera setups. For that, we supplement the ORB-SLAM2 \cite{mur2017orb} state-of-the-art visual SLAM algorithm with a preprocessing module that consists of the classification neural network. The aim of the network is to determine the quality of the input images, that is, a conjectural number of ORB features \cite{rublee2011orb} that the ORB-SLAM2 algorithm will base its estimation on.

For evaluation of the proposed approach, we have collected a dataset using the HermesBot mobile platform \cite{protasov2021cnn}, equipped with six rolling-shutter cameras facing in different directions. The dataset composes of typical indoor scenes, including open spaces, corridors, plain walls, empty rooms, as well various kinds of static and dynamic obstacles like chairs, tables, moving people, etc. Thanks to the camera position setup and the chosen dataset location, each video sequence consists of both feature-rich and feature-poor images, depicting typical data collected on an indoor ground robot trajectory. Two of the sequences include Ground Truth collected using a 3D LiDAR.

This dataset is then used for the training of the neural network to predict whether the image has enough ``good'' ORB features, that are correctly matched between consecutive frames features, to be transferred to the SLAM pipeline. Thus, the localization is performed using the camera with the best resulting feature number. We evaluate the accuracy of neural net classification, validate whether choosing the best camera approach maintains the number of good features sufficient for correct SLAM operation, and compare the metrics of computational efficiency of our approach and other possible multi-camera schemes.



\section{System Overview}\label{section:overview}
\subsection{Hardware and Software Architecture}
For the proposed study, we have used the HermesBot platform developed for the research in the area of autonomous ground robotics, depicted in Fig. \ref{hermes_render}. The platform consists of a base with six motor-wheels equipped with encoders, a metal frame with a number of sensors installed on it, including, among others, six rolling-shutter RasPi NoIR V2.1 8-megapixel cameras, and a Velodyne VLP-16 LiDAR used in the proposed study.
The cameras are installed on both sides of the robot, facing sideways and at the angle of 45 degrees to the sides. The area covered by all the cameras is depicted in Fig. \ref{fig:hermes_fov}. Prior to the dataset collection, all cameras were calibrated using a chessboard pattern, and the transformation matrices for each camera were calculated based on their location on the robot.


The platform is controlled using a tandem of two computational units: an Intel NUC computer featuring an Intel Core i7 processor for general control and data handling, and an Nvidia Jetson Xavier module for image processing. All sensors and modules are connected using the ROS framework. Each camera is connected to a separate Raspberry Pi 3 model B+ single-board computer for image acquisition and preprocessing. On the computers, the images are published into ROS topics via a Gigabit Ethernet switch.
The neural network module is installed on both the NUC and the NVIDIA Xavier computing modules, with the former utilizing CPU for computation, and the latter using GPU. This was done in order to evaluate the performance of the proposed approach on both CPU and GPU and compare it with the traditional ORB-SLAM2 pipeline, and the ORB feature extractor and matcher, that can only operate on a CPU.




The scheme of our system's software architecture and general data processing pipeline is depicted in Fig. \ref{fig:pipeline_architecture}. AGX Xavier, Intel NUC and cameras share a common ROS environment. Each camera posts its frames into a corresponding ROS image topic. AGX Xavier listens to new images and fits a batch of them into the neural network. Each batch consists of 6 images, one image per camera. The network predicts the probability for an image to have enough features for the correct operation of ORB-SLAM2 \cite{mur2017orb}, the visual SLAM algorithm chosen for the study of multi-camera operation. The image to be transferred on ORB-SLAM2 input is chosen based on the network prediction.

\subsection{Dataset}



For the training and evaluation of the proposed approach, we have collected a dataset of visual data in Skoltech campus. The dataset consists of three sequences recorded in different areas of the building and of different length: \textit{Sequence 1} – 3825 frames on each of 6 cameras, \textit{Sequence 2} – 6376 frames, \textit{Sequence 3} – 2546 frames. In all sequences, the cameras are recording the data at 20 FPS. During collection, the data is preprocessed on Raspberry Pi computers: the images are resized to the resolution of 640×480 and fed into ROS topics. The data is recorded in rosbag format, with every image provided with a unified timestamp in the robot's time, and then converted into PNG. 


\textit{Sequence 2} and \textit{Sequence 3} have intersections in the robot position, so, to avoid overfitting, they were used for the network training, and \textit{Sequence 1} was used for evaluation.

\begin{figure} [!t] 
\vspace{0.5em}
\begin{center}
\includegraphics[width=0.45\textwidth]{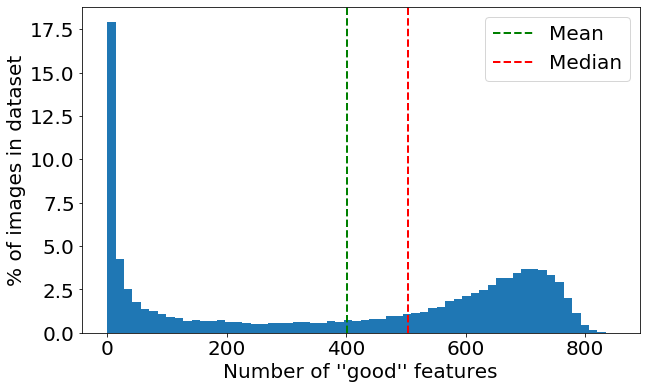}
\caption{The histogram of ``good" features on the dataset images including \textit{Sequences 1, 2} and \textit{3}.
}
\vspace{-1.8em}
\label{fig:hist}
\end{center}
\end{figure}

\subsection{Dataset Processing}
After the dataset collection, all frames were processed using the ORB feature extractor and matcher, and only the ``good'' features that were matched with the subsequent frame were used for network training and evaluation. 

To determine whether a feature is ``good'', we evaluated the matched keypoint descriptors by compiling a fundamental matrix for each pair of consecutive images. This allowed us to train and evaluate the neural network only on keypoints that were successfully matched with the subsequent frame, and thus, ensured the correct operation of ORB-SLAM2.

An example of the ``good'' features histogram is depicted in Fig. \ref{fig:hist}. The histogram shows that the number of features varies from 0 to 1000, with mean around 300 features per image and median at about 400. We used a threshold to divide images into 2 classes: images with number of “good” features is higher than the threshold are labeled as “good” ones, otherwise as bad ones. Based on the data distribution, we chose the threshold as 350 features.

\subsection{Network Architecture}
We used the pre-trained ResNet-18 \cite{he2016deep} convolutional part as the backbone, and after its average pooling layer, we added a fully connected head. The fully connected part consists of two hidden layers with 512 neurons with ReLu activation and one final layer with 2 neurons with SoftMax activation. Each neuron returns the network confidence on the image relating to the corresponding class. The network is trained to the classification task using Adam optimizer with learning rate 0.001, beta1 0.9 and beta2 0.999. We trained both the pretrained ResNet-18 backbone and the head during 8 epochs with batch size of 128 images on our dataset.

\section{Experiments}


\begin{figure*} [!t] 
\begin{center}
\includegraphics[width=16.4cm]{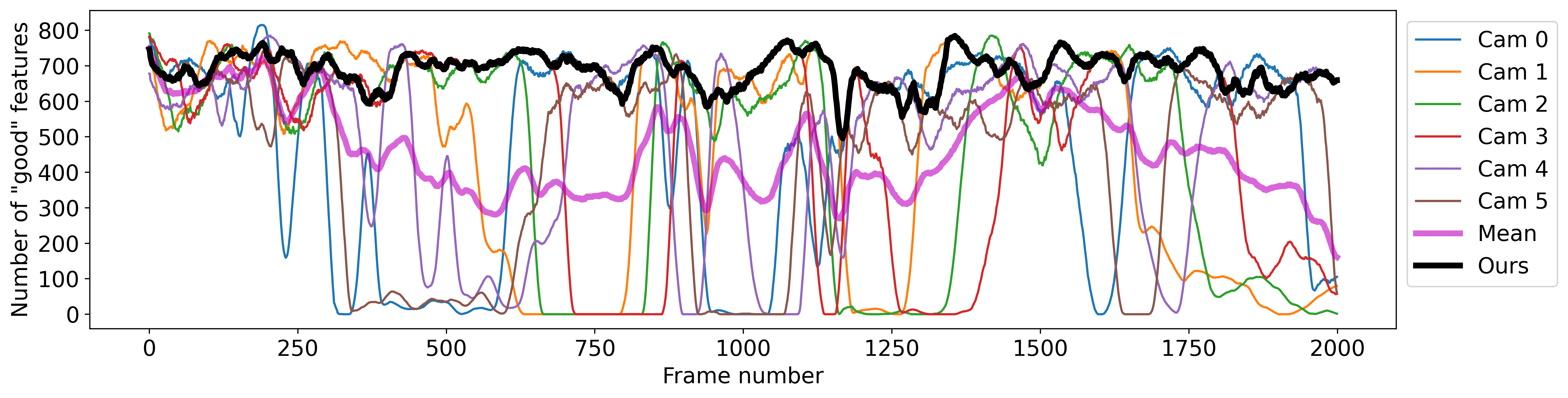}
\vspace{-0.5em}
\caption{The number of ``good'' ORB features on consecutive frames of each camera, and the mean value for all cameras, \textit{Sequence 1}. The number of ``good'' features using the neural network predictions is depicted in black.}
\vspace{-2em}
\label{fig:exp2}
\end{center}
\end{figure*}
In order to evaluate the proposed approach, three sets of experiments were carried out. The first set of experiments was devoted to the evaluation of the neural network on its accuracy on estimating the images with the highest number of ``good'' features, that is, the features that can be matched with the subsequent frame. The second set evaluated the overall number of features if the ``switch-to-best'' approach is used utilizing the neural network, compared to the number of ``good'' features on single cameras. The third set of experiments was aimed at calculating the performance of the proposed approach compared to traditional visual SLAM, and estimating the same features using the original ORB feature extractor and matcher. The performance of the neural network on GPU and CPU was evaluated as well.


\subsection{Neural Network Accuracy Evaluation}
We trained the neural networks on Sequences 2 and 3 in order to validate their performance on Sequence 1, that does not intersect its trajectory with the sequences for training. To evaluate the efficiency of the classifier, we used the classification accuracy score, that is the amount of true positive and negative predictions with respect to the overall number of neural network predictions, and F1 score, that is a harmonic mean of precision and recall, where the best value is 1 and worst score is 0. 

Table \ref{table:exp1} demonstrates both the high level of accuracy and F1 score on each camera of validation Sequence 1. Among all cameras, the average precision is 0.926, and the average F1 score is 0.924. Thus, the proposed architecture is able to robustly predict the camera frames that are most suitable to be fed into the ORB-SLAM algorithm and the high number of potentially matched ORB features is 90\% guaranteed according to the results in Table \ref{table:exp1}. 

\begin{table}[htb]
\vspace{0.5em}
\caption{Performance Metrics of NN}
\begin{tabular}{|l|l|l|l|l|l|l|}
\hline
\backslashbox{Metic}{Cam}
         & 0     & 1     & 2     & 3     & 4     & 5     \\ \hline
Accuracy & 0.908 & 0.944 & 0.927 & 0.92  & 0.924 & 0.931 \\ \hline
F1 Score & 0.901 & 0.942 & 0.92  & 0.916 & 0.932 & 0.935  \\ \hline
\end{tabular}
\label{table:exp1}
\vspace{-1.5em}
\end{table}

\subsection{Validation on Collected Dataset}

The trained neural network was launched on each camera image of Sequence 1, and the information about the amount of ``good'' potentially matched features was obtained. The best camera with the highest values of ``good'' features at each timestamp was chosen based on the neural network predictions. So, we have obtained the sequence of camera numbers with the best quality to use in the SLAM pipeline during the operation. Launching the extraction and matching of ORB features on the obtained sequence with images from cameras chosen by the neural network, gave us promising results, shown in Fig. \ref{fig:exp2}. By switching to the camera that was predicted to have the highest number of ``good'' features, we obtained a new sequence, depicted ``Ours'' in Fig. \ref{fig:exp2}, with the average of 691 features per frame. Another generated sequence in Fig. \ref{fig:exp2} is ``Mean'' that is the mean number of ``good'' features on all six cameras, averaging at only 386. The results confirm that the proposed approach works well and the sufficient number of features can be maintained during the operation using the camera switches based on the neural network predictions.

\subsection{Performance Evaluation}

In order to validate the performance of our system, we evaluate the speed of our neural network on the CPU i7-6700K 4.00GHz and GPU GTX 1070. The experiment has two parts: the first one was devoted to find out whether the processing of ORB extractor + matcher is less efficient than the NN approach of figuring out the camera feature ``goodness''. The experiments have shown that one iteration over six images takes:
\begin{itemize}
    \item 120 ms using the traditional ORB matcher on a CPU;
    \item 10 ms using the neural network on a CPU;
    \item 2.7 ms using the neural network on a GPU.
\end{itemize}
According to the experimental results, our approach is significantly faster than ORB feature matching: approximately 12 times on the CPU and approximately 50 times on the GPU. The results also show that the use of matching algorithms taken from ORB-SLAM2 to assess the quality of features will significantly load the system and decrease the effective rate of the pipeline, while our approach increases the computational complexity of the pipeline insignificantly, even if it is launched on a CPU, that is less optimal for neural networks.

In the course of the second part of the experiment, we compare the performance of ORB-SLAM2 running on one camera and on 6 cameras, ORB-SLAM2 using our neural network on current and the 5 remaining cameras and one ORB-SLAM2 with an estimate of the number of orb feature matches on current and 5 remaining cameras. ORB-SLAM2 and matching processes were launched on CPU throughout the whole experiment, since its architecture makes it unable to utilize GPU. Table \ref{table:exp3_2_new} shows that the use of a neural network in the system will increase the ORB-SLAM2 computational time by only 20 percent if running on a CPU and by 5 percent if running on the GPU. The use of a standard feature matching algorithm increases the processing time almost 3.5 times. Thus, on a mobile platform with limited computational resources, it would not allow ORB-SLAM2 to operate in real time. On the testing hardware, the performance of even two ORB-SLAM2 pipelines with our neural network has achieved near real-time operation. Therefore, it may be possible to use the two best resulting cameras instead of one, further increasing the robustness and reliability of the system.
\begin{table}[h]
\vspace{-1 em}
\caption{Performance of Different Setups of ORB-SLAM2 Pipeline }
\vspace{-1em}
\begin{center}
\begin{tabular}{|c|cc|l}
\hline
\multirow{2}{*}{Method}  & \multicolumn{2}{c|}{Time of one iteration}  \\ \cline{2-3}
                         & \multicolumn{1}{c|}{Without GPU} & With GPU \\ \hline
1xORB-SLAM                & \multicolumn{1}{c|}{48 ms}       & -         \\ \hline

6xORB-SLAM               & \multicolumn{1}{c|}{285 ms}      & -         \\ \hline
1xORB-SLAM+6xORB-matcher & \multicolumn{1}{c|}{168 ms}      & -         \\ \hline
1xORB-SLAM+6xNN (ours)          & \multicolumn{1}{c|}{58 ms}       & 50.7 ms   \\ \hline
2xORB-SLAM+6xNN          & \multicolumn{1}{c|}{106 ms}      & 98.7 ms   \\ \hline
\end{tabular}
\label{table:exp3_2_new}
\end{center}
\vspace{-1.5em}
\end{table}





\section{Conclusion}

In this work, we presented an approach to improving the computational efficiency and robustness of visual SLAM  algorithms on mobile robots with multi-camera setups. This was achieved using a neural network based on ResNet18.
This network makes a prediction about whether the image contains enough  ORB features that can be matched with subsequent frames. 
Extensive experiments have shown that the proposed solution is optimal both in terms of image estimation quality and performance.
The average number of ``good'' features on the data proposed by the neural network amounted to 688, as opposed to an average of 386 among all frames on all cameras. This value, and the absence of feature level drop, make it possible to ensure the reliable operation of the ORB-SLAM2 algorithm even if images from most cameras do not contain a sufficient amount of keypoints.
The computational complexity increase was only 5\% when adding our solution to the SLAM pipeline on 6 cameras. 
The iteration time using the proposed neural network was only 2.7 ms when using the GPU and 10 ms on the CPU. This proves that our solution can increase the robustness of visual SLAM in multi-camera setups without a significant decrease in performance.



\section{Discussion and Future Work}
Our future work will be devoted to implementing the camera switching, pipeline into existing visual SLAM approaches, as well as ensuring the smoothness of transitions between them. The SLAM initialization time will be taken into account as well. The performance of the proposed approach will be further compared with state-of-the-art visual SLAM approaches, and depicted in further publications.

Also, our future work will focus on testing the proposed approach on various hardware configurations and platforms. The neural network can be extended for use with multiple types of cameras, including fisheye and global shutter cameras. This would allow evaluating it on other robotic platforms developed at Skoltech, including the Ultrabot disinfection robot \cite{perminov2021ultrabot, mikhailovskiy2021ultrabot}, a platform for plant disease detection \cite{karpyshev2021autonomous}, a robot for shopping rooms \cite{petrovsky2020customer} and even drone-based and multi-robot solutions\cite{kalinov2021impedance, kalinov2020warevision, kalinov2021warevr}, where the computational efficiency issue is the most critical.

\section*{Acknowledgements}
The reported study was funded by RFBR and CNRS according to the research project No. 21-58-15006.




\bibliographystyle{IEEEtran} 
\bibliography{case_karpyshev2022}

\end{document}